\documentclass{elsarticle}  % Comment this line out
\usepackage{amsmath} % assumes amsmath package installed
\usepackage{amssymb}  % assumes amsmath package installed
\usepackage{mathtools}
\usepackage{svg}
\usepackage{graphics}
\usepackage{subfig}
\usepackage{paralist}
\usepackage[normalem]{ulem}

\journal{Neurocomputing}

\begin{document}

\begin{frontmatter}

\title{Grounding the Experience of a Visual Field through Sensorimotor Contingencies}

\author{Alban Laflaqui\`ere\corref{cor1}}
\ead{alaflaquiere@aldebaran.com}
\cortext[cor1]{Corresponding author}
\address{AI Lab, SoftBank Robotics Europe, 43 rue du Col. Pierre Avia, 75015 Paris, France}

\begin{abstract}
Artificial perception is traditionally handled by hand-designing task specific algorithms. However, a truly autonomous robot should develop perceptive abilities on its own, by interacting with its environment, and adapting to new situations. The sensorimotor contingencies theory proposes to ground the development of those perceptive abilities in the way the agent can actively transform its sensory inputs. We propose a sensorimotor approach, inspired by this theory, in which the agent explores the world and discovers its properties by capturing the sensorimotor regularities they induce. This work presents an application of this approach to the discovery of a so-called visual field as the set of regularities that a visual sensor imposes on a naive agent's experience. A formalism is proposed to describe how those regularities can be captured in a sensorimotor predictive model. Finally, the approach is evaluated on a simulated system coarsely inspired from the human retina.
\end{abstract}

\begin{keyword}
autonomous systems \sep developmental robotics \sep sensorimotor contingencies \sep predictive processing \sep sensorimotor learning \sep human-like vision 
\end{keyword}

\end{frontmatter}

%%%%%%%%%%%%%%%%%%%%%%%%%%%%%%%%%%%%%%%%%%%%%%%%%%%%%%%%%%%%%%%%%%%%%%%%%%%%%%%%%%%%%%%%%%%%%%%%%%
%-------------------------------------------------------------------------------------------------
%%%%%%%%%%%%%%%%%%%%%%%%%%%%%%%%%%%%%%%%%%%%%%%%%%%%%%%%%%%%%%%%%%%%%%%%%%%%%%%%%%%%%%%%%%%%%%%%%%

\section{Introduction}
\label{sec:Introduction}

Autonomy in robotics relies on sensory data processing to capture information about the world and adapt to it. Although the influence of machine learning has been growing more important in the last decades, traditional approaches to this problem of data processing involve significant manual design from engineers that build the robot. Consequently the resulting techniques for artificial perception appear rigid and constrained for tractability. Each of these specialized algorithms is applicable to only a small set of tasks, with potentially limiting inbuilt biases from the designer. While acceptable for well-defined processes, such as industrial manufacturing, the potential need for a large degree of human involvement makes such methods inadequate as a source of long-term autonomy in a robot. Instead, an autonomous robot must be able to cope with the complexity of its world, build its own way to perceive it and adapt to its variations.
%It must in particular learn to control the \emph{interface} with this world that its body (motors and sensors) constitutes \cite{The interface Theory of perception, Hohwy (blanket stuff}.

To address this issue, the field of developmental robotics takes inspiration from biological and cognitive development in children \cite{cangelosi2015developmental}. It proposes that an agent learns to interact with its environment, autonomously and on an ontogenic timescale. Without prior knowledge, a naive robot must learn the structure of its own body, of its environment, and how the two interact. 
In this context, \emph{perception} is a prerequisite to developing more advanced cognitive abilities that allow a rich interaction with the environment. Yet, the emergence of this fundamental capacity, traditionally hand-coded in the system, poses a challenge: What is \textit{perception} for a naive agent in which manually pre-defined features and labels are replaced by a flow of uninterpreted sensorimotor data?

Sensorimotor contingencies theory (SMCT) attempts to answer this question \cite{o2001sensorimotor} by fundamentally re-defining perception: \emph{perception is the mastery of regularities in the way actions transform sensory inputs}. It suggests that a naive agent can actively explore its environment, extract regularities that the world imposes on its sensorimotor flow, and later identify those regularities when interacting with the environment in order to perceive it. Those regularities, or \emph{contingencies}, are the ground on which the agent can build perceptive abilities.
Moreover this active account of perception naturally links actions to perception, meaning that the agent intrinsically knows what it could \emph{do} with any perceived feature \cite{seth2014predictive}.
Despite its philosophical aspect, the SMCT is based on experimental results. Among other things, it elegantly accounts for instance for sensory substitution. Those are experiences in which a subject is provided with information from one modality (e.g. vision) through another modality's pathway (e.g. skin or ears) \cite{bach1969vision,auvray2007learning}. The theory naturally encompasses such a phenomenon as it defines perception as based on structure in the sensorimotor flow instead of properties of the pathway it takes. Such a possibility leads us to consider artificial ``plug-and-play" agents that could be equipped with new sensors, and would discover how to perceive with them by learning to master the associated sensorimotor flows.
%It also explains more common experiences of extending one's feeling of embodiment, for instance when driving a car or playing a video game.

The SMCT has been relatively slow to spread in the robotics community, partly because of the complete overhaul it induces in the field of artificial perception. To date, the approach has been applied to model the acquisition of perceptive concepts such as space \cite{laflaquiere2015learning, terekhov2013space}, colors \cite{philipona2006color,witzel2015}, environments \cite{laflaquiere2015grounding}, and objects \cite{laflaquiere2015objects}. Primarily, these works characterize properties of the external world explored by the agent. However, a naive agent's body is also part of the unknown world it has to discover. It contributes, like the structure of the environment, to shaping the regularities the agent experiences in its sensorimotor flow. As such, properties of the agent's body should also be captured through sensorimotor contingencies.
In this paper, we address the problem of capturing properties of sensors plugged on a naive agent, and in particular properties of the \emph{visual field} generated by visual sensors.
The experience of visual field encapsulates the set of regularities describing how visual features are encoded differently by various parts of the sensor, as well as how they shift on it due to motion. This fact is particularly striking when considering heterogeneous visual sensors, like the human retina, for which visual features are encoded by significantly different cell patterns depending on where they land on the retina.
This discrepancy between our stable subjective experience of visual features and their actual variable sensory encoding has already been brought forward in the paper introducing the SMCT \cite{o2001sensorimotor}. Yet, only recently has it led to further inquiries with the development of psycho-motor experiments \cite{herwig2014predicting}. Their results suggest that the brain learns the correspondence between the different sensory patterns that encode the same visual feature on different parts of the retina, and the motor commands (ocular saccades) that transform one into the other. By exploring artificial visual setups where two distinct visual features are consistently associated before and after a saccade, it is possible to alter previously learned correspondences. This artificial interaction with the world leads to a modification of the subjective perceptive experience of visual features, even in adult subjects.

The work presented in this paper proposes a computational model inspired by this perceptive phenomenon. Nonetheless it also fits into a more general endeavor to develop a computational model for the autonomous learning of sensorimotor regularities \cite{laflaquiere2015grounding, maye2013, seth2014predictive}, the lack of which has been the second reason of the slow spread of SMCT.
The formalism converges towards the hierarchical building of a predictive model of sensorimotor experiences \cite{laflaquiere2015grounding}. This approach is in line with recent developments in neuroscience, which describe the brain as a predictive machine \cite{friston10,clark2013whatever,hohwy2014self}. By learning to predict future sensory outcomes of its actions, the agent estimates latent causes of its experience and progressively extends the control from its motor component to its sensory component. The work presented in this paper will focus on letting a naive agent discover the sensorimotor regularities that define the \emph{visual field} associated with a visual sensor.
The next section presents a formalization of the problem and describes a computational model to address it. A simulation is then introduced in Sec.\ref{sec:Simulation} to illustrate the approach. The results are analyzed in Sec.~\ref{sec:Results} in light of previous works in the sensorimotor approach of perception. Finally, limitations and potential future extensions of the model are discussed~in the last section.

%%%%%%%%%%%%%%%%%%%%%%%%%%%%%%%%%%%%%%%%%%%%%%%%%%%%%%%%%%%%%%%%%%%%%%%%%%%%%%%%%%%%%%%%%%%%%%%%%%
%-------------------------------------------------------------------------------------------------
%%%%%%%%%%%%%%%%%%%%%%%%%%%%%%%%%%%%%%%%%%%%%%%%%%%%%%%%%%%%%%%%%%%%%%%%%%%%%%%%%%%%%%%%%%%%%%%%%%

\section{Problem Formulation}
\label{sec:Formulation}

In this section we present the problem a naive agent is facing when discovering the sensorimotor structure induced by its visual sensor. We describe the regularities that underlie the experience of a \emph{visual field}. Then we propose a computational formalism to process the agent's sensorimotor flow and detail how it can capture those regularities.

\subsection{Experiencing a visual field}
\label{sec:Experiencing a visual field}

\begin{figure}[t]
\includegraphics[width=0.8\linewidth]{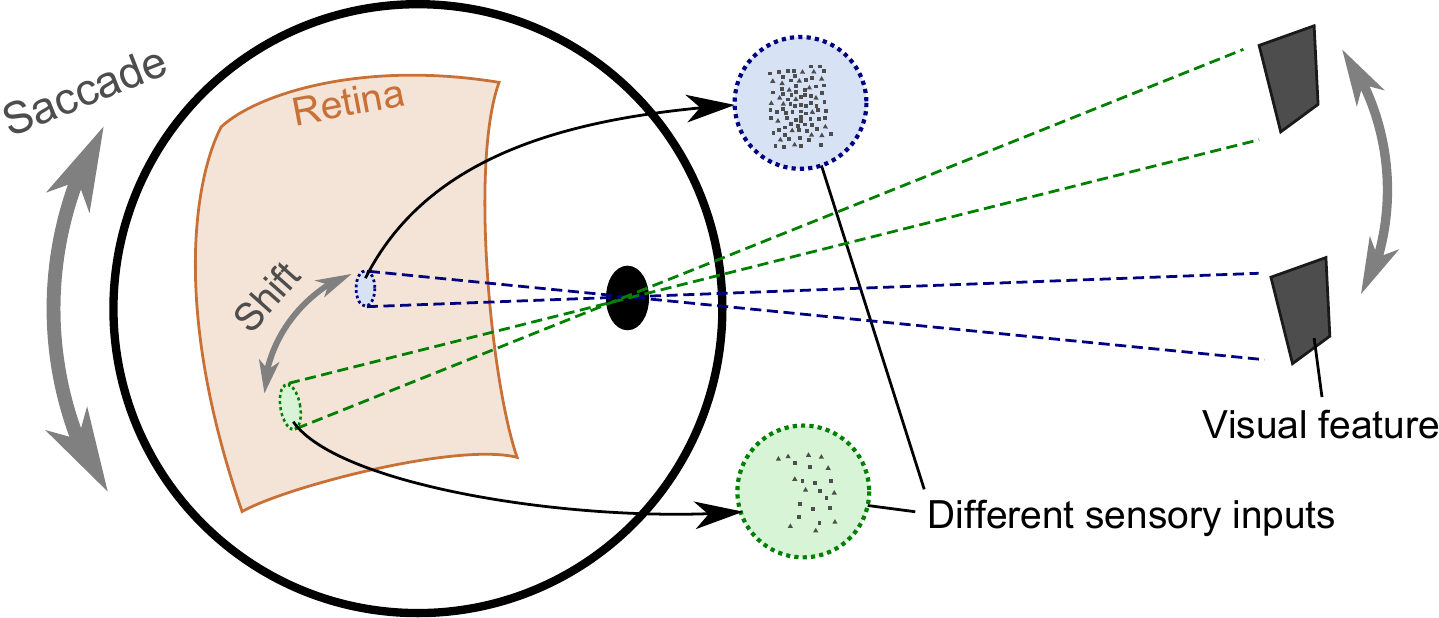}
\centering
\caption{The heterogeneous structure of the retina implies that visual features are encoded by different sensory inputs depending on where they are in the visual field. These different sensory inputs can be experienced by saccading with the eye.}
\label{fig:Retina_principle}
\end{figure}

This work focuses on agents equipped with a visual-like sensor: an array of sensels collecting information from a part of the environment, where a sensel is the basic element of a sensor array (e.g. pixels in a camera, or rods and cones in our retina).
In this work, we use the term \textit{visual feature} to refer to the visual information received from a small part of the environment. Contrarily to computer vision literature where visual features are the internal outcome of some sensory processing, the term here describes the (partial) state of the external environment. Conversely, we use the term \textit{sensory inputs} to refer to the information generated when visual features are projected on the sensor and transformed into an encoded signal accessible to the agent (see Fig.~\ref{fig:Retina_principle}).

Depending on where it is present in the visual field, a visual feature can be projected onto different parts of the sensor array. It can thus be encoded by different sensory inputs. Such a claim does not appear obvious when considering a camera because it is usually assumed that the sensory encoding is translation-invariant: physical properties of the sensor are such that a visual feature generates the same sensory input regardless of where it is encoded in the array. This is for instance an implicit hypothesis in Convolutional Neural Networks, a class of algorithms that prove to be very efficient in visual scene analysis \cite{krizhevsky2012imagenet}. It also indirectly assumes that the later unit that processes sensory inputs does know the spatial organization of sensels and can switch on the fly between different groups of neighboring sensels.

Yet, such a property is far from evident for a biological system like our visual cortex. This fact appears even less realistic when taking into account the heterogeneity of the human retina \cite{records1979physiology}. As underlined in \cite{o2001sensorimotor}, the way visual features are encoded changes significantly across the retina, due to its physiological properties. Yet, our subjective experience of visual features is that they are stable across the whole field of view. The sensorimotor point of view on such a phenomenon claims that the brain learns to associate the different sensory inputs corresponding to the same visual feature by actively exploring visual scenes. This hypothesis has been recently strengthened by psycho-motor experiments in which those associations were artificially altered \cite{herwig2014predicting}.

According to SMCT, the very mastery of those sensorimotor associations participates in the experience of seeing. More precisely, by focusing on regularities induced by the physical structure of the sensor, one can describe those that give rise the experience of having a \emph{visual field}:
\begin{itemize}
\item Sets of different sensory inputs encode the same visual features on different parts of the sensor.
\item Motor commands can transform sensory inputs into another one encoding the same visual feature.
\end{itemize}
Those two statements describe that visual features shift on the retina and that their encoding changes when the agent moves its sensor (see Fig.~\ref{fig:Retina_principle}).
It is important to notice that the only way for a naive agent to discover such properties is to actively explore visual scenes. They could not be extracted through a passive sensory processing, like the ones usually proposed in unsupervised contexts \cite{bengio2013representation}. For instance, the different sensory inputs related to a single visual feature would not necessarily share the same statistical properties or lie close to one another in the sensory space, especially if the sensor is significantly heterogeneous. Additionally, passively extracted knowledge wouldn't be directly useful to a naive agent as it wouldn't know how to actively transform its sensory state to eventually reach a goal state (for instance, move the sensor to bring a visual feature in a given part of the visual field). Yet, the association of sensory inputs seems intimately linked to the ability to perform visual tasks such as search and recognition, as demonstrated in \cite{herwig2014predicting}.
%It also contributes to the subjective experience of uniform acuity in our field of view, even when peripheral vision is actually very coarse.

We claim that a naive agent has to explore its environment and learn the sensorimotor regularities induced by its visual sensor. In line with the Predicting Processing approach \cite{clark2013whatever}, which describes the brain as a predictive machine, we propose to capture those regularities in a predictive model described hereunder.

\subsection{Formalization and learning of the predictive model}
\label{sec:Formalization and learning of the predictive model}

The phenomenon described in Sec.~\ref{sec:Experiencing a visual field} suggests that the sensory experience generated by the visual sensor can be studied locally. Taking inspiration from biology, and in particular the structure of the human retina, we treat the array as a cluster of \emph{receptive fields} \cite{hubel1968receptive}. Each receptive field is made of numerous neighboring sensels that cover a limited part of the whole array, all with the same physical extent. No additional constraint is assumed and the different receptive fields may have different properties: e.g. the number of sensels, their spatial arrangement, or their excitation function. Receptive fields encode independently the visual feature they receive from the environment. For the naive agent, each receptive field initially appears to be an independent sensor generating its own sensory input. 

Formally, we define the \emph{sensory input} generated by a receptive field as a multivariate random variable $\mathbf{S}^a$ that can take values: 
\begin{equation}
\mathbf{s}^a_i = [s_{i,1},\dots,s_{i,d^a}],
\end{equation} 
where $i$ denotes the index of the sensory input, $a$ denotes the receptive field, $s_{i,k}$ is the individual sensation provided by the $k^{th}$ sensel in receptive field $a$, and $d^a$ is the number of sensels in this receptive field.
The agent is able to move its visual sensor using saccades, analogous to human eye movements. Formally, we denote the saccadic motor commands as a multivariate random variable $\mathbf{M}$ that can take values:
\begin{equation}
\mathbf{m_q} = [m_{q,1},\dots, m_{q,d^m}],
\end{equation}
where $q$ denotes the index of the motor command, $m_{q,k}$ is the individual command sent to the $k^{th}$ motor moving the sensor, and $d^m$ is the number of those motors. No specific superscript is needed for the motor command as all receptive fields move together in a rigid manner when the sensor is moved.
Like with the human eye, we assume that saccades are fast enough that the state of the environment can be considered constant during the execution of most of the motor commands. Moreover, sensory inputs are supposed to be generated instantaneously by each receptive field, before and after a saccade.

According to our sensorimotor approach, the naive agent needs to explore actively the world to capture regularities underlying its sensorimotor experiences. In line with a Bayesian description  of the brain \cite{knill2004bayesian}, we propose to capture those regularities by letting the agent build a predictive model of its sensorimotor experiences. Note that we're interested in how the agent can actively transform its experience, which means that it has to model sensorimotor transitions. A similar modeling of sensorimotor transitions has already been proposed in the literature, although often intimely linked to a reinforcement learning framework \cite{sutton2011horde,mugan2012autonomous,ghadirzadeh2016self}.
More precisely here, the agent should estimate the probability:
\begin{equation}
P\big(\mathbf{S}^b(t+1)\:|\:\mathbf{S}^a(t)=\mathbf{s}^a_i,\mathbf{M}(t)=\mathbf{m}_q \big),
%\forall b, \forall a, \forall i, \forall q
\end{equation}
corresponding to the conditional post-saccadic distribution of sensory input $\mathbf{S}^b$ in receptive field $b$, given that a sensory input $\mathbf{s}^a_i$ was experienced in receptive field $a$ before the execution of the saccadic motor command $\mathbf{m}_q$. For the sake of simplicity, this probability is later denoted:
\begin{equation}
P(\mathbf{S}^b\:|\:\mathbf{s}^a_i,\mathbf{m}_q).
\end{equation}
The agent can estimate this distribution by collecting multiple instances of sensorimotor transitions,
\begin{equation}
(\mathbf{s}^a_i,\mathbf{m}_q \rightarrow \mathbf{s}^b_j),
\end{equation}
and statistically estimate the distribution over $j$.
This model is relatively simple but notice that multiple ones have to be estimated in parallel as the sensor is made of numerous receptive fields.

The physical embedding of the sensor in the world does not allow all possible sensorimotor transitions. Those implicit constraints should appear as regularities in the predictive model: transitions imposed by the sensor and the world should have high probabilities while ``forbidden" ones should be improbable.
In particular, the definition of a visual field proposed in Sec.~\ref{sec:Experiencing a visual field} can now be further formalized by introducing the following set:
\begin{equation}
\mathcal{G}_{\mathbf{s}^a_i} = \{\mathbf{s}^b_j \:|\: \exists \mathbf{m}_q : P(\mathbf{s}^b_j\:|\:\mathbf{s}^a_i,\mathbf{m}_q) \geq \epsilon\}
\label{eq:similarity_set}
\end{equation}
where $\epsilon$ is a threshold defined to distinguish significant probabilities from unsignificant ones. The set $\mathcal{G}_{\mathbf{s}^a_i}$ gathers all sensory inputs $\mathbf{s}^b_j$ that the agent can transition to with a probability greater than $\epsilon$ from input $\mathbf{s}^a_i$ by performing a motor command $\mathbf{m}_q$. In our approach, the probability of active sensorimotor transitions can be seen as a similarity measure between the two involved sensory inputs $\mathbf{s}^a_i$ and $\mathbf{s}^b_j$ \cite{laflaquiere2015grounding}. For a fixed significance threshold $\epsilon$, $\mathcal{G}_{\mathbf{s}^a_i}$ can thus be seen as the set of all sensory inputs $\mathbf{s}^b_j$ considered similar to $\mathbf{s}^a_i$ by the agent.

In the next section, we introduce a simulated system to evaluate our approach on two different kinds of environments. The predictive models estimated by the agent after exploring those environments will later be analyzed in detail in Sec.~\ref{sec:Results}.

%%%%%%%%%%%%%%%%%%%%%%%%%%%%%%%%%%%%%%%%%%%%%%%%%%%%%%%%%%%%%%%%%%%%%%%%%%%%%%%%%%%%%%%%%%%%%%%%%%
%-------------------------------------------------------------------------------------------------
%%%%%%%%%%%%%%%%%%%%%%%%%%%%%%%%%%%%%%%%%%%%%%%%%%%%%%%%%%%%%%%%%%%%%%%%%%%%%%%%%%%%%%%%%%%%%%%%%%

\section{Simulation}
\label{sec:Simulation}

\begin{figure}[t]
\includegraphics[width=1\linewidth]{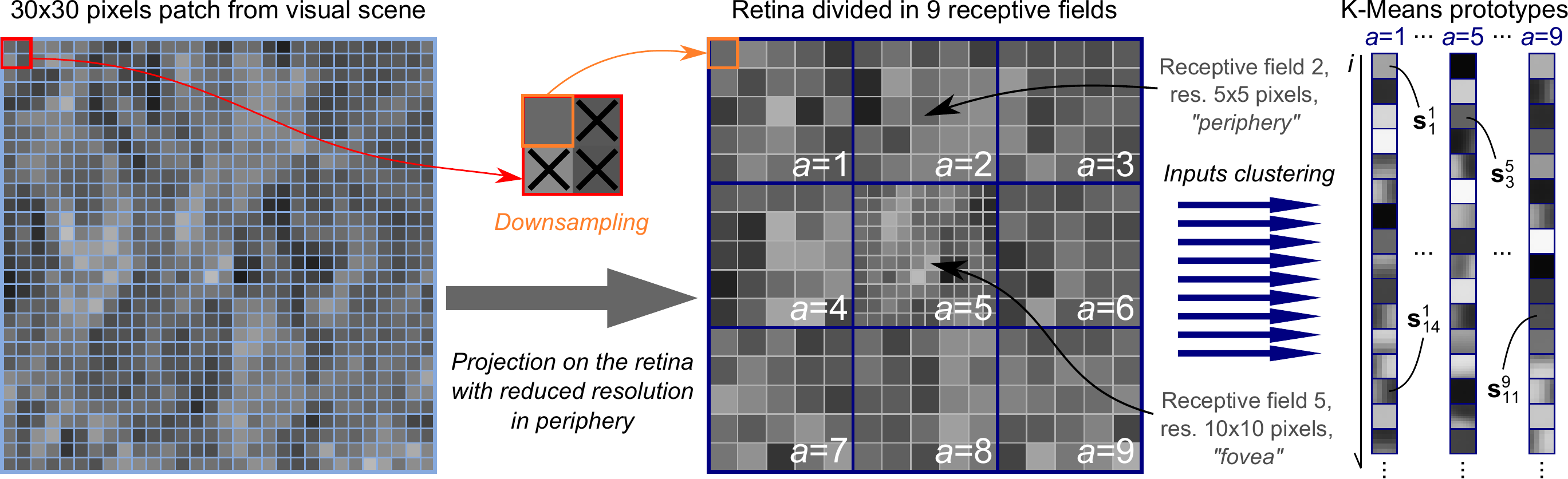}
\caption{The simulated visual sensor is coarsely inspired by the retina structure. \textit{On the left}, the sensor has a field of view limited to a $30\times 30$ pixels patch which captures information from the environment. \textit{In the center}, this initial patch is divided into smaller $10\times 10$ ones corresponding to the $9$ receptive fields. Simultaneously, the resolution of all $8$ peripheral receptive fields is halved both in height and width to form the final inputs accessible to the agent. \textit{On the right}, independent K-Means are run for the different receptive fields to generate sensory prototypes and cluster later sensory inputs.}
\label{fig:Simulated_retina}
\end{figure}

\begin{figure}[t]
\includegraphics[width=1\linewidth]{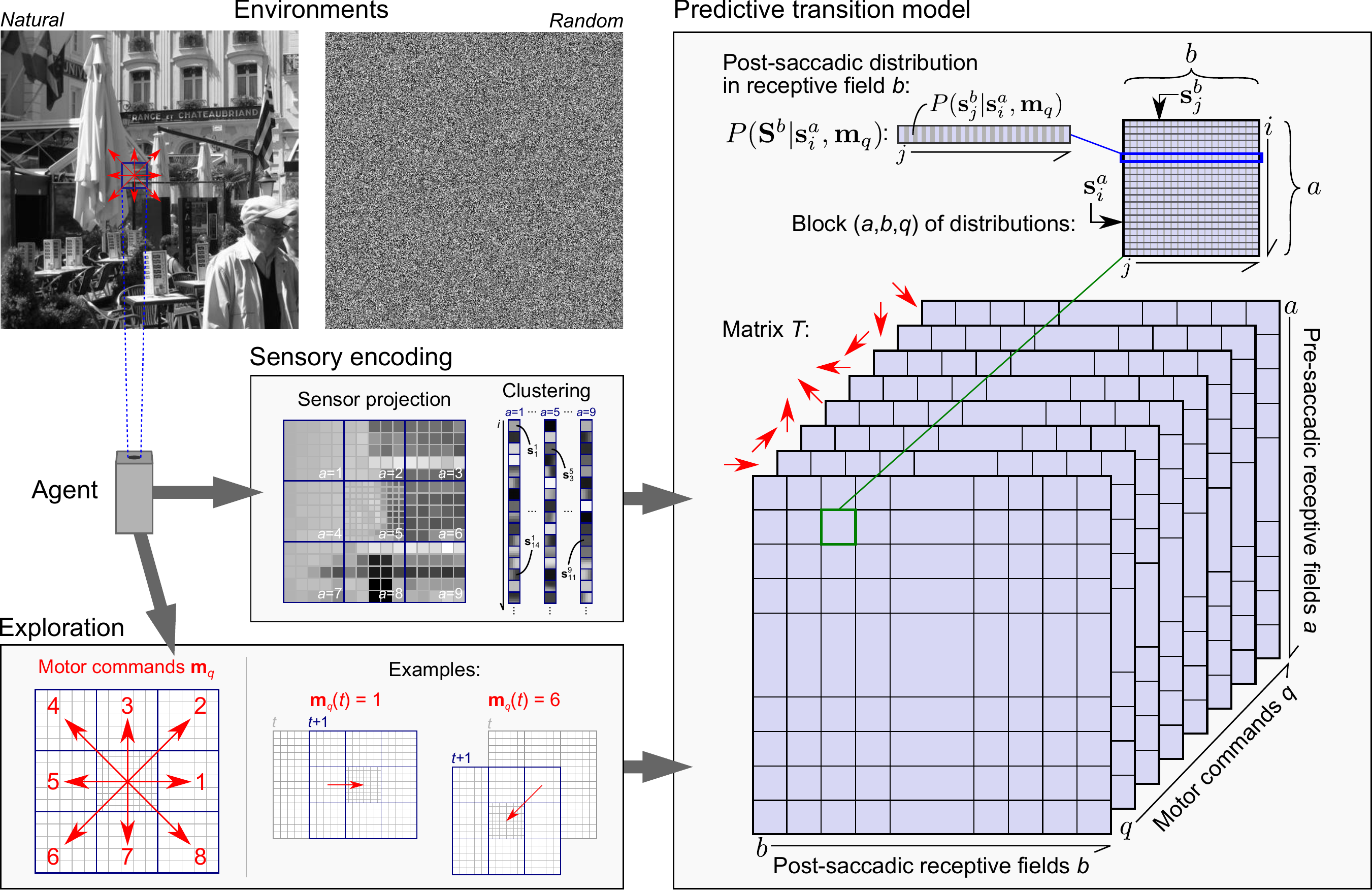}
\caption{The agent interacts with its environment through its visual sensor. \textit{On the top left}, two kinds of environments, random and natural images, are considered and explored during two independent runs of the simulation. \textit{On the center left}, the narrow patch of information captured by the sensor is encoded by $9$ independent sensory states. \textit{On the bottom left}, the agent actively explores the visual environment by randomly performing $8$ saccades that shift the locations of receptive fields. \textit{On the right}, the agent builds a predictive model of the sensorimotor transitions it experiences. It is made up of multiple distributions over post-saccadic states $\mathbf{s}^b_j$ estimated for each pre-saccadic receptive field $a$ and state $i$, each post-saccadic receptive field $b$ and each motor command $q$. Individual distributions related to the same values of $a$ and $b$ are grouped to form 'blocks', which are themselves grouped according to the value of $q$ to form the $3$D matrix $T$.}
\label{fig:Simulation}
\end{figure}

A simple system is simulated in order to illustrate the approach. In this section, we describe the agent's sensory and motor systems, the different kinds of visual scenes considered during exploration, as well as the algorithmic details of the predictive model estimation.

\subsection{Simulated agent}
\label{sec:agent}

As illustrated in Fig.~\ref{fig:Simulated_retina} and \ref{fig:Simulation}, the simulated system intends to coarsely capture the kind of interaction a moving eye has with its environment. The agent is a translatable camera exploring a two-dimensional visual environment. Its field of view is limited to a narrow $30\times 30$ pixels square.
As such, it captures only a small portion of what a human retina would capture in a visual scene. The sensor must thus be seen as a simple model of what happens in a small part of an actual retina.
The field of view is divided into $9$ receptive fields of size $10\times 10$ pixels. However, the resolution of each peripheral receptive field is artificially lowered to imitate the coarser sensory encoding in the peripheral retina compared to the central fovea. The reduced resolution is obtained by downsampling the original $10\times 10$ inputs: only one column in every two and one row in every two are considered to form a $5\times 5$ pixels input (see Fig.~\ref{fig:Simulated_retina}). Pixels discarded this way are considered nonexistent in the sensor and do not interfere in any way in the remaining pixels excitation (for instance by applying a local average or a max-pooling criterion). A fourth of the original information is thus accessible to the agent in the peripheral layer of the retina. Finally, the sensory inputs $\mathbf{s}^a_i$ generated by the different receptive fields are respectively of size $d^a=100$ for the central receptive field ($a=5$), and $d^a=25$ for peripheral ones ($a\neq5$).

In order to limit the simulation complexity, we consider that each receptive field $a$ can take only a restricted set of $N^a$ prototypical sensory inputs $\mathbf{s}^a_i, i\in [1,N^a]$, called sensory states. 
The constant $N^a$ is arbitrarily set in relation to the associated sensory space dimensionality: $N^a=50$ for the central (foveal) receptive field ($a=5$), and $N^{a}=20$ for the peripheral ones ($a\neq5$). The prototypical sensory states $\mathbf{s}^a_i$ are determined in a data-driven way by collecting a large number of sensory inputs and applying a clustering algorithm. Pragmatically, we let the agent randomly explore multiple environments and collect $10^6$ sensory inputs per receptive field. Independent K-Means \cite{lloyd1982least} are then run on the data collected in each receptive field to generate the prototypical sensory states. These latter are thus potentially different between receptive fields, even with the same resolution. Such data clustering can be seen as the analogous encoding that visual data undergoes just after the retina \cite{marcelja1980mathematical}.
In later exploration, sensory inputs received in a receptive field are simply encoded by the closest prototypical sensory state $\mathbf{s}^a_i$ (winner-takes-all strategy).

The agent can translate in the visual scene to sample different parts of its environment. Similarly to sensory inputs, movements are discretized into a limited set of $Q=8$ saccades $\mathbf{m}_q$. They correspond to all translations of the retina such that the central receptive field shifts to the pre-saccadic location of one of the 8 peripheral receptive fields (see Fig.~\ref{fig:Simulation}). Those saccades have been purposely chosen so that visual features entirely shift between receptive fields during movements of the sensor, which will facilitate the results analysis.

\subsection{The environment}
\label{sec:environments}

The environment corresponds to visual scenes, simple images that the agent can explore one at a time.
%The visual scenes that the agent can explore are two-dimensional images.
Two different types of images are considered in the simulation (see Fig.~\ref{fig:Simulation}):
\begin{itemize}
\item \textit{Random}: gray-scale images of size $1024\times 1024$ pixels generated by randomly drawing the integer value of each pixel independently from a uniform distribution $\text{U}(0,255)$.
\item \textit{Natural}: images taken from a standard RGB database \cite{mit-saliency-benchmark} and converted to gray-scale pixel values.
\end{itemize}
The simulation can be run independently on each type of visual environment.
The random images are intended to test the system in the absence of any environmental structure. It corresponds to an optimal setting in which only the sensor structure constrains the agent's sensorimotor experience. On the contrary, natural images are used to evaluate the approach in a more realistic setting.
Note that we do not want the system to over-fit its model to a specific visual scene. For each type of environment, $100$ different scenes are thus generated/drawn for the agent to successively explore during the simulation.
As explained in Sec.~\ref{sec:Formalization and learning of the predictive model}, we assume that saccades are fast enough so that the environment can be considered static while most sensorimotor transitions are experienced.
More precisely, we hypothesize that the probability of having the environment change (draw a new image) during a saccade  is significantly lower than the probability of executing a saccade (the ratio is arbitrarily set to $1$ to $10^4$).
%Visual scenes are thus changed in-between two saccades to comply with this assumption.
Such a dynamic appears reasonable when considering the speed of a human saccade. This constraint may nonetheless be loosened by assuming a greater ratio.
%that the environment is \emph{statistically} static during saccades.
Implicitly this assumption means that if this temporal stability hypothesis were to be broken and the environment would always change during saccades, the agent could not discover enough regularities to structure an experience of visual field. This kind of assumption has also been recently described as a requirement for robotic systems \cite{jonschkowski2015learning}.

\subsection{Building the predictive model}
\label{sec:Building the predictive model}

We want the agent to model its interaction with the world by building a predictive model of the sensorimotor transitions it can experience.
As claimed by the sensorimotor approach described previously, the agent needs to actively explore its environment in order to build this model. To this end, the agent is arbitrarily placed at the center of its environment (center position in the first image it explores) and let to explore by randomly executing $10^6$ successive motor commands. The visual scene is changed after every $10^4$ saccades, so that the agent successively interacts with all the $100$ scenes. This random exploration policy is natural for a naive agent that has no a priori knowledge about itself nor the environment. It can be seen as analogous to body babbling observed in babies.
Each saccade generates an elementary sensorimotor transition experience $(\mathbf{s}^a_i,\mathbf{m}_q\rightarrow \mathbf{s}^b_j)$ for each $a$ and $b$. Given that the sensor is divided into $9$ receptive fields, this sums up to $81$ sensorimotor transitions for each saccade.
The agent estimates each discrete distribution $P(\mathbf{S}^b\:|\:\mathbf{s}^a_i,\mathbf{m}_q)$ based on the collected data by simply building a normalized histogram of outputs $\mathbf{s}^b_j$ for each triplet $(a,i,q)$. Intuitively, the robot estimates in which state receptive field $b$ will statistically be if receptive $a$ is in state $i$ and it performs the motor command $q$.

For visualization, all atomic predictive models are stored in a 3D matrix $T$ (see Fig.~\ref{fig:Simulation}). Each distribution $P(\mathbf{S}^b\:|\:\mathbf{s}^a_i,\mathbf{m}_q)$ is stored as a row vector obtained by concatenating the individual probabilities $P(\mathbf{s}^b_j\:|\:\mathbf{s}^a_i,\mathbf{m}_q), \forall j \in [1,N^b]$. All the possible distributions generated by the different values of $a$, $i$, $b$, and $q$ are then concatenated to form $T$ such that: its rows correspond to the pre-saccadic states $i$ for the different values of $a$, its columns correspond to the post-saccadic states $j$ for the different values of $b$, and its pages correspond to the executed saccadic motor command $q$.
This way, the matrix $T$ can be interpreted by ``blocks", which are 2D matrices corresponding to the predictive structure between pairs of receptive fields $(a,b)$ for a given saccade $q$. For this reason, we will later refer to these blocks using the triplet $(a,b,q)$ (see Fig.~\ref{fig:Simulation}).

In the next section, we analyze the predictive model built by the agent while exploring the two kinds of environments. In particular, we will focus on describing the sensorimotor structure captured by the naive agent and that defines properties of the visual field generated by its sensor.

%%%%%%%%%%%%%%%%%%%%%%%%%%%%%%%%%%%%%%%%%%%%%%%%%%%%%%%%%%%%%%%%%%%%%%%%%%%%%%%%%%%%%%%%%%%%%%%%%%
%-------------------------------------------------------------------------------------------------
%%%%%%%%%%%%%%%%%%%%%%%%%%%%%%%%%%%%%%%%%%%%%%%%%%%%%%%%%%%%%%%%%%%%%%%%%%%%%%%%%%%%%%%%%%%%%%%%%%

\section{Results}
\label{sec:Results}

As claimed by the sensorimotor approach described in Sec.~\ref{sec:Formulation}, physical properties of the agent's visual sensor and its related visual field constrain the sensorimotor interaction it has with the world. Those constraints should appear as a highly predictable structure in the predictive model built by the agent. Hereunder, we identify and analyze this structure for the two kinds of environments considered in the simulation.

\begin{figure}[!t]
\includegraphics[width=1\linewidth]{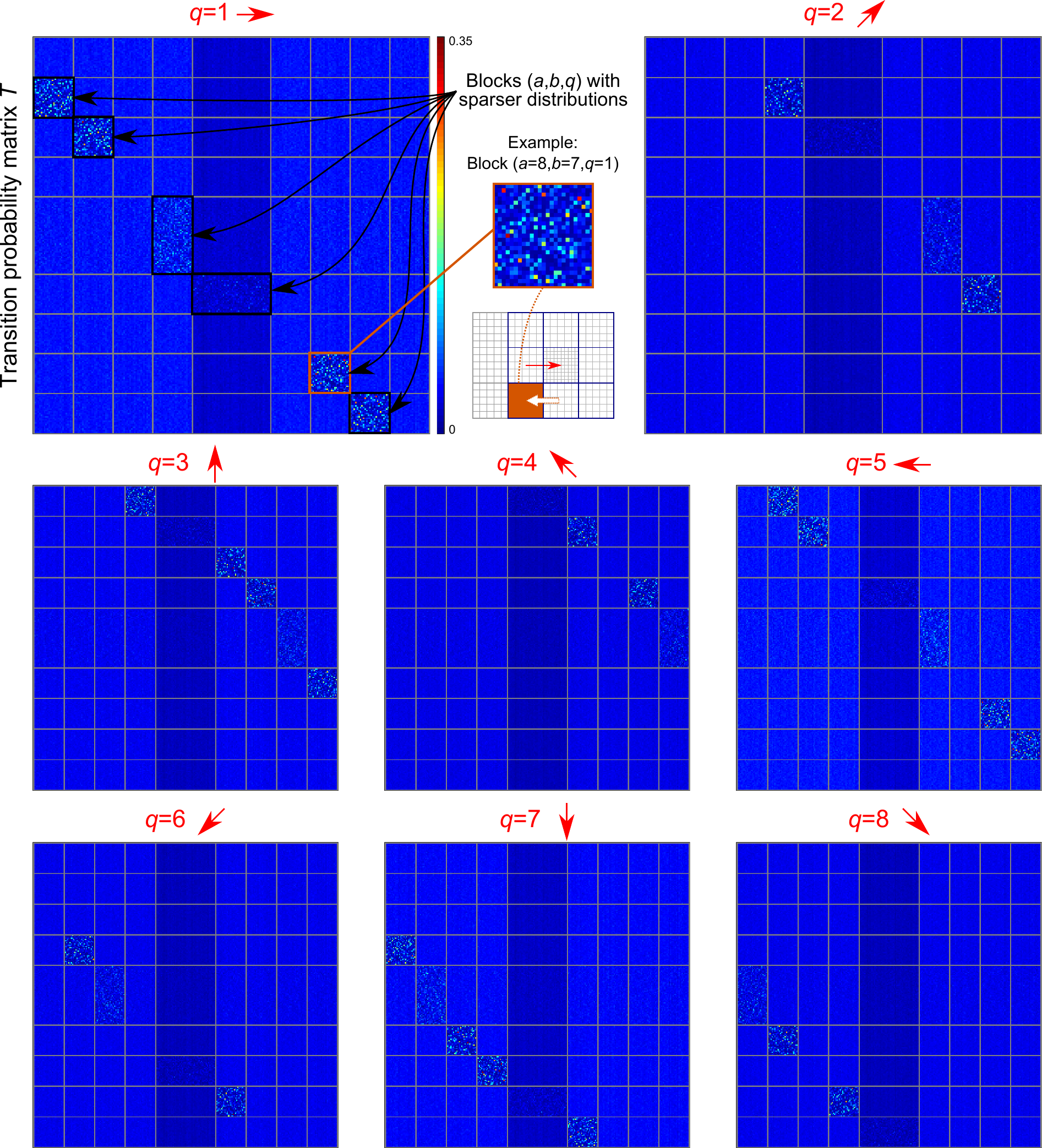}
\caption{The $3$D matrix $T$ estimated by exploring random visual scenes and displayed one page at a time for each motor command $q$. Distributions in the whole matrix are generally uniform. However, for each saccade, a few blocks $(a,b)$ display sparser distributions. This is for instance the case of block $(a=8,b=7)$ for the saccade $q=1$. Looking at the sensor's structure, this higher predictability can be explained by the shift of visual features between corresponding receptive fields when the saccade is performed. (best seen in color)}
\label{fig:Matrices_noise}
\end{figure}

\subsection{Random images}
\label{sec:Random images}

The matrix $T$ built by the agent after exploring random images is presented in Fig.~\ref{fig:Matrices_noise}. Although unrealistic, this environmental setup is optimal to study how the agent does capture the structure induced by its visual sensor. Indeed no environmental structure is expected to influence the learning process.

Transitions probabilities in the whole matrix $T$ are generally very low. Such a result is not surprising in a random environment as it indicates that the agent cannot accurately predict future sensory states $\mathbf{s}^b_j$ based on current sensory states $\mathbf{s}^a_i$ and the saccade $\mathbf{m}_q$. Globally, each block $(a,b,q)$ displays close to uniform distributions over $j$ for the different pre-saccadic states $i$. Those blocks thus appear as almost uniform blue patches in Fig.~\ref{fig:Matrices_noise}. The central column of blocks that corresponds to predicting the sensory state of the foveal receptive field ($b=5$) also appears darker. This is simply because those distributions $P(\mathbf{S}^5\:|\:\mathbf{s}^a_i,\mathbf{m}_q)$ are spread across a larger number ($N^b=50$) of sensory states than for peripheral receptive fields ($N^b=20$).

Yet a few blocks $(a,b,q)$ appear to exhibit a different structure due to sparser distributions (see highlighted blocks in Fig.~\ref{fig:Matrices_noise}). In other terms, for each saccade $q$, there exist pairs of receptive fields $(a,b)$ for which knowing $\mathbf{s}^a_i$ helps predicting $\mathbf{s}^b_j$.
When checking the values $(a,b,q)$ associated with those blocks and which receptive fields/saccades they correspond to, one can notice that they actually capture the shift of visual features on the sensor during saccades. As an example, the first ``structured" block encountered in the matrix $T$ is $(a=2,b=1,q=1)$. When looking at the actual sensor structure in Figures~\ref{fig:Simulated_retina}\ and~\ref{fig:Simulation}, one can verify that any visual feature projected in the receptive field $a=2$ gets transferred to receptive field $b=1$ when the saccade $q=1$ moves the sensor to the right.
Without specifying in advance the structure of the sensor, we can see that its properties have been implicitly captured in the predictive sensorimotor model built by the agent. Based on this model, the agent can determine relations between its receptive fields (which initially appear as unknown independent sensors) and its motor commands. It can know - or more precisely did estimate - that some sets of sensory states exist, where members of the set can be transformed into one another by sending motor commands (see Eq.~\ref{eq:similarity_set}).
%Formally, we can define those sets $\mathcal{G}_{\mathbf{s}^a_i}$ as:
%\begin{equation}
%\mathcal{G}_{\mathbf{s}^a_i} = \{\mathbf{s}^b_j \:|\: \exists \mathbf{m}_q : P(\mathbf{s}^b_j\:|\:\mathbf{s}^a_i,\mathbf{m}_q) \geq \epsilon\}
%\end{equation}
%with $\epsilon$ a significance threshold that will be discussed hereafter.
From an external point of view, those sets correspond to the different sensory encodings of visual features projected on different parts of the sensor. The model of sensorimotor transitions built by the agent has thus intrinsically captured properties of the visual field generated by the sensor. Importantly enough, those properties are not internally represented as ungrounded symbolic information but directly in a sensorimotor model that can be used by the agent to interact with the world and perform visual tasks \cite{herwig2014predicting}.
An example of such a visual task would be to look into the estimated predictive model for the saccade that would transform, with the highest probability, a current sensory input into another desired one. Such an algorithm has been recently proposed in \cite{laflaquiere2016ICDL}.

\textit{Influence of the model:}
The maximal values of $P(\mathbf{s}^b_j\:|\:\mathbf{s}^a_i,\mathbf{m}_q)$ are relatively low, even in structured blocks where predictability should theoretically be very high.
This phenomenon is due to the simplicity of the predictive model we proposed. As a reminder, sensory inputs of each receptive field have been clustered using the K-Means algorithm. This method minimizes distances between samples and their closest prototype, and form convex clusters. However, those clusters are not necessarily ``aligned" between the receptive fields. This means that all the data falling into a cluster $i$ of one receptive field can fall at the intersection of multiple clusters $j$ when projected into the sensory space of another receptive field. The effect is particularly marked when exploring random images: sensory samples can be scattered in the whole sensory space, which consequently generates a high variance in the positions of K-Means prototypes between the receptive fields. The consequence of this misalignment for the predictive model is that each sensory state can be transformed into a few other states in the associated receptive field when the sensor saccades (see highlighted block in Fig.~\ref{fig:Matrices_noise}). Maximal predictability is thus relatively low, even in couple blocks.

\textit{Information transfer:}
Despite those limitations, we were able to visually distinguish a predictive structure in coupled blocks $(a,b,q)$. However, in order to evaluate this structure more formally, we propose to introduce a measure derived from information theory. For each block $(a,b,q)$ we compute the normalized conditional entropy $H(a\,|\,b,q)$ as follows:
\begin{equation}
H(a\,|\,b,q) = -\sum_{i,j}\frac{P(\mathbf{s}^b_j,\mathbf{s}^a_i\:|\:\mathbf{m}_q)}{\log N^b} \log \bigg( \frac{P(\mathbf{s}^b_j,\mathbf{s}^a_i\:|\:\mathbf{m}_q)}{P(\mathbf{s}^a_i\:|\:\mathbf{m}_q)} \bigg)
\end{equation}
Entropy is a measure of uncertainty in the statistical model. Intuitively $H(a\,|\,b,q)$ measures the unpredictability of sensory input $\mathbf{S}^b$ in receptive field $b$ given the input $\mathbf{S}^a$ in receptive field $a$ and the saccade $\mathbf{m}_q$. Consequently, $H(a\,|\,b,q)$ should be significantly lower in blocks coupled by the structure of the sensor than in others.

As shown in Fig.~\ref{fig:Entropies}, this is indeed the case as coupled blocks appear darker than their counterparts. This measure validates the fact that, even if the sensory clusters are misaligned in the different sensory spaces, a transfer of information does occur between coupled receptive fields when saccades are performed. This transfer is directly caused by the physical structure of the sensor.
One can also notice that coupled blocks involving the foveal receptive fields display a slightly higher entropy than when they only involve peripheral ones (for instance, blocks $(a=5,b=4,q=1)$ and $(a=6,b=5,q=1)$ for the first motor command). This is the result of multiple causes that reduce the quality of the predictive mapping between a peripheral sensory space and the foveal one: the dimension of the foveal sensory space is higher than in periphery, the number of foveal prototypes is also higher than in periphery, and the loss of resolution (see Sec.~\ref{sec:agent}) induces a lower predictability between peripheral and foveal sensory states.

%the mapping between sensory states, which does not preserve the data metrics. Specifically, this mapping corresponds to the averaging of groups of 4 pixels into (meta-)pixels described in \ref{sec:agent}. Clusters defined by K-Means in the foveal sensory space are distorted when projected in the peripheral sensory space via this mapping. Consequently, the distorted projection of a foveal cluster does not correspond to a cluster that K-Means could generate in the peripheral sensory space; instead it overlaps with many of them. The resulting one-to-many clusters mapping leads to a lower predictability.

%This one-to-many clusters mapping explains why the predictive model lacks accuracy compared to a setting where clusters would be perfectly aligned.

\subsection{Natural images}
\label{sec:Natural images}

\begin{figure}[!t]
\includegraphics[width=1\linewidth]{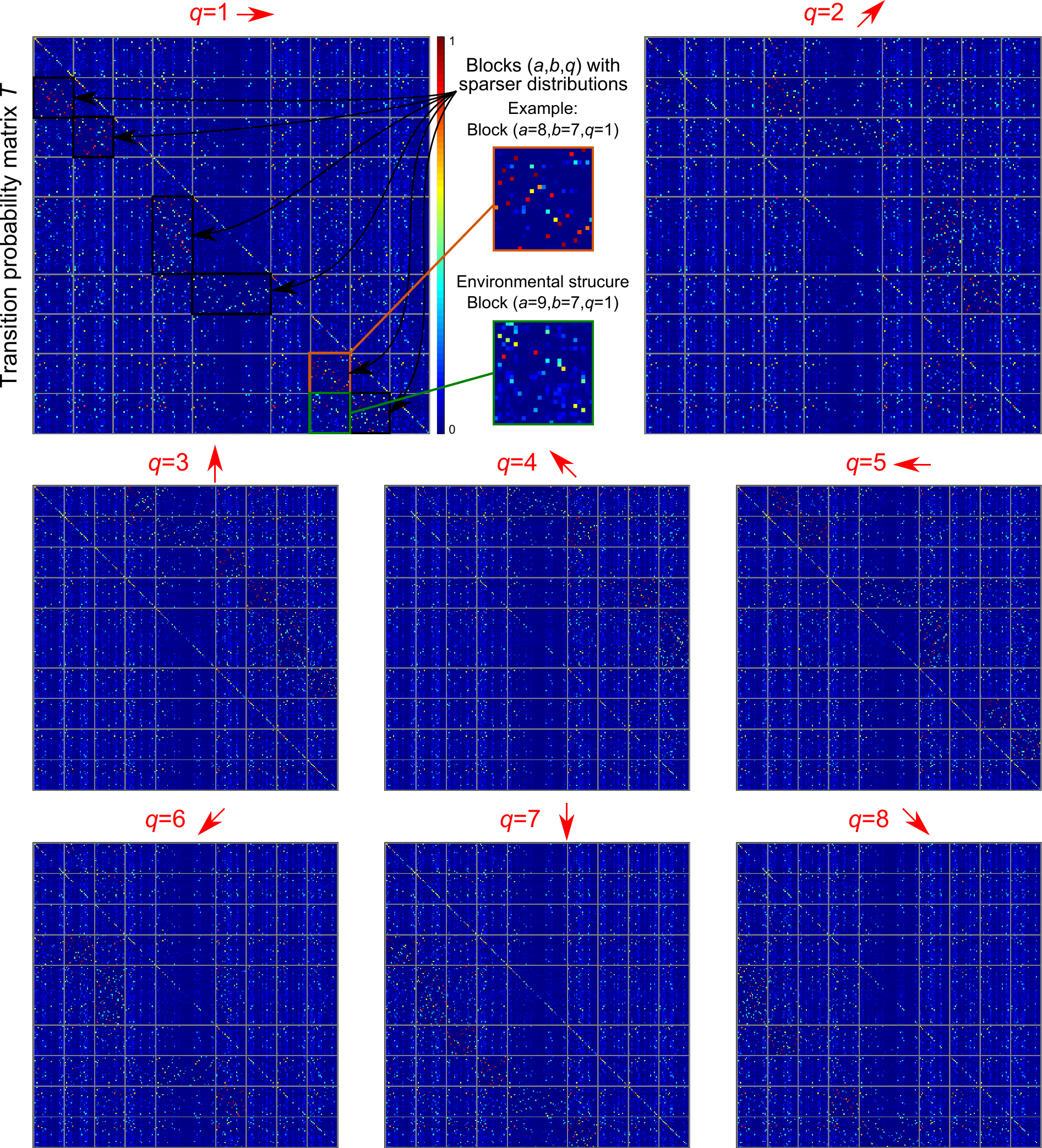}
\caption{The $3$D matrix $T$ estimated by exploring natural visual scenes and displayed one page at a time for each motor command $q$. Distributions in the whole matrix are sparser than when exploring random images because environmental structure is captured in the predictive model. However, same blocks $(a,b,q)$ as in the previous simulation display higher predictability. They correspond to receptive fields coupled by the physical structure of the sensor and between which visual features shift when a saccade is performed. (best seen in color)}
\label{fig:Matrices_natural}
\end{figure}

The matrix $T$ built by the agent after exploring natural images is presented in Fig.~\ref{fig:Matrices_natural}. This setup is proposed to evaluate the approach in a more realistic environment containing structure.

\begin{figure}[!t]
\includegraphics[width=1\linewidth]{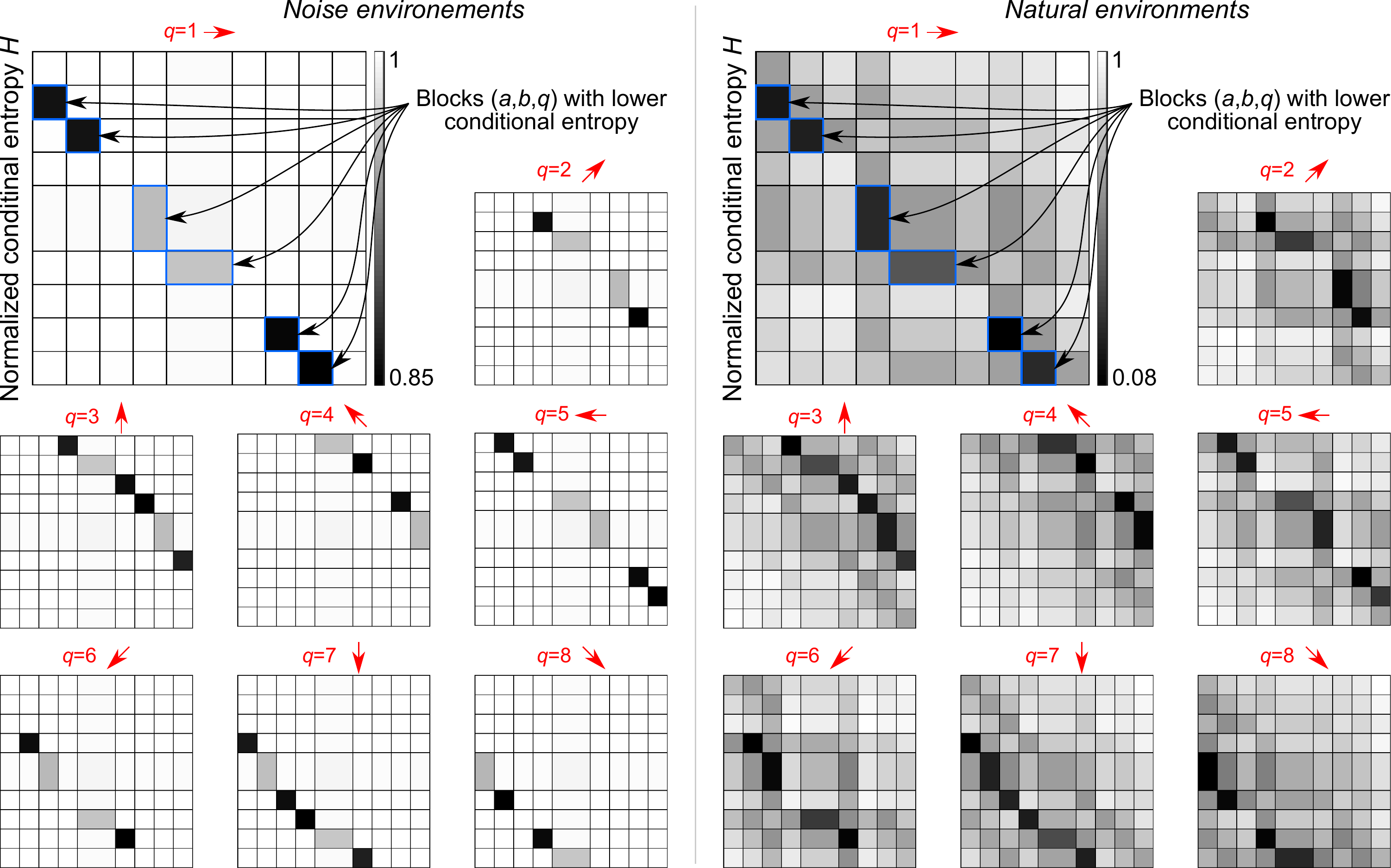}
\caption{Normalized conditional entropies $H(a\,|\,b,q)$ computed on blocks $(a,b,q)$ and displayed so as to keep the shape as matrix $T$. For both kinds of visual scenes, $H(a\,|\,b,q)$ has a significantly lower value in the coupled blocks already identified in matrices $T$. This measure formally demonstrated that information is transferred between some receptive fields when a saccade is performed. From an external point of view, this transfer is due to the physical structure of the sensor.}
\label{fig:Entropies}
\end{figure}

Overall the estimated model appears to contain more predictive structure than when the agent was exploring random environments. The different blocks in the matrix $T$ generally display sparser distributions. This visual analysis is confirmed by the measure of conditional entropies $H(a\,|\,b,q)$ that are significantly lower than in the previous scenario (see Fig.~\ref{fig:Entropies}).
Such a result is to be expected when the agent explores natural images. Those visual scenes indeed have some intrinsic structure which implies that, locally, visual features can help predict other features with significant accuracy. For instance, surfaces in the world tend to keep the same appearance and receiving a uniform black visual feature in a receptive field allows to predict that neighboring receptive fields also probably observe the same visual feature. The same way, one can predict that all those receptive fields will most probably receive a black uniform feature regardless of the saccade the agent performs.
This environmental structure constrains the sensorimotor experience of the agent and naturally gets captured in the sensorimotor predictive model it estimates.

Nonetheless, apart from this environmental structure that appears scattered everywhere in the matrix $T$, one can also notice that some blocks $(a,b,q)$ display even sparser distributions. They correspond to the same blocks that were already identified in the previous scenario (for instance, blocks $(a=2,b=4,q=2)$ or $(a=8,b=4,q=8)$). As shown in Fig.~\ref{fig:Entropies}, their conditional entropy measures $H(a\,|\,b,q)$ are significantly lower than their counterparts, which confirms the fact that information gets largely transferred between those paired receptive fields when the corresponding saccade is performed. Once again, the constraints imposed by the sensor's structure have been captured by the agent and define its visual field experience.

Notice however that the structure in those blocks is much sparser than in the previous simulation (see Fig.~\ref{fig:Matrices_natural} compared to Fig.~\ref{fig:Matrices_noise}). In other words, in coupled blocks $(a,b,q)$, knowing the pre-saccadic sensory state $\mathbf{s}^a_i$ allows to predict with a very high probability the post-saccadic sensory state $\mathbf{s}^b_j$, given $\mathbf{m}_q$. The reason of such a good predictive mapping is that visual data provided by natural images is not scattered in the whole sensory space of each receptive field like it was the case for the randomly generated data. They instead tend to naturally cluster in subparts of the sensory space. The independent K-Means performed in the different receptive fields' sensory spaces thus tend to converge towards similar clusters. Those clusters are thus relatively well aligned, which means that data falling into one cluster in a receptive field almost entirely falls into a single cluster when projected into another receptive field. This alignment allows the predictive model to reach very high probabilities for those transitions.

Nonetheless, two exceptions can be identified. They both correspond to transitions involving the fovea (see Fig.~\ref{fig:Matrices_natural}).
The first one is when the foveal sensory state predicts the post-saccadic sensory state of a coupled peripheral receptive field (for instance block $(a=5,b=4,q=1)$, or more generally any coupled block with $a=5$). In those blocks, some distributions $P(\mathbf{S}^b\:|\:\mathbf{s}^a_i,\mathbf{m}_q)$ are very sparse, with a single peak (appearing red in the color code), while some display multiple peaks of lower probability. This phenomenon is due to the different quantizations of the foveal sensory space, with $N^a=50$ prototypes, and the peripheral ones, with $N^a=20$ prototypes. Intuitively, the $50$ clusters, once projected into a peripheral sensory space, cover a space that is otherwise covered by $20$ bigger clusters. Some former clusters thus get entirely included into a bigger one while others end up partially overlapping multiple adjacent bigger clusters. The first case leads to perfect predictability while the second case leads to sparser distributions. This explains the particular predictive structure observed in those blocks.
The second exception is when a peripheral sensory state predicts the post-saccadic sensory state of the fovea (for instance block $(a=6,b=5,q=1)$, or more generally any coupled block with $b=5$). In those blocks, distributions are not as sparse as in blocks involving only peripheral receptive fields. Consequently, they also display intermediary values in the conditional entropy matrix (see Fig.~\ref{fig:Entropies}). This result can be explained by two factors. First, the difference of number of prototypes between the two sensory spaces has the inverse effect of the previous case. The $20$ big peripheral clusters overlap many smaller clusters when projected in the foveal sensory space. An optimal configuration where a big cluster would be included entirely in a smaller one thus never happens. This leads to lower predictability of the post-saccadic foveal state. The second factor is not related to the computational model used in the simulation but to the sensor properties. The difference of resolution between the peripheral and foveal receptive fields is such that the peripheral encoding of a visual field does not contain enough information to determine precisely its foveal encoding. This ambiguity does set an upper limit on the transition probabilities in those blocks. Despite those limitations, those blocks remain the ones with the lowest conditional entropy $H(a\,|\,b,q)$ compared to blocks in the same column $b=5$ for each saccade $q$. This impossibility to optimally predict the foveal sensory state would thus not prevent the agent to associate those pairs of receptive fields when trying to predict next foveal experiences.

\begin{figure}[!t]
\centering
\includegraphics[width=0.85\linewidth]{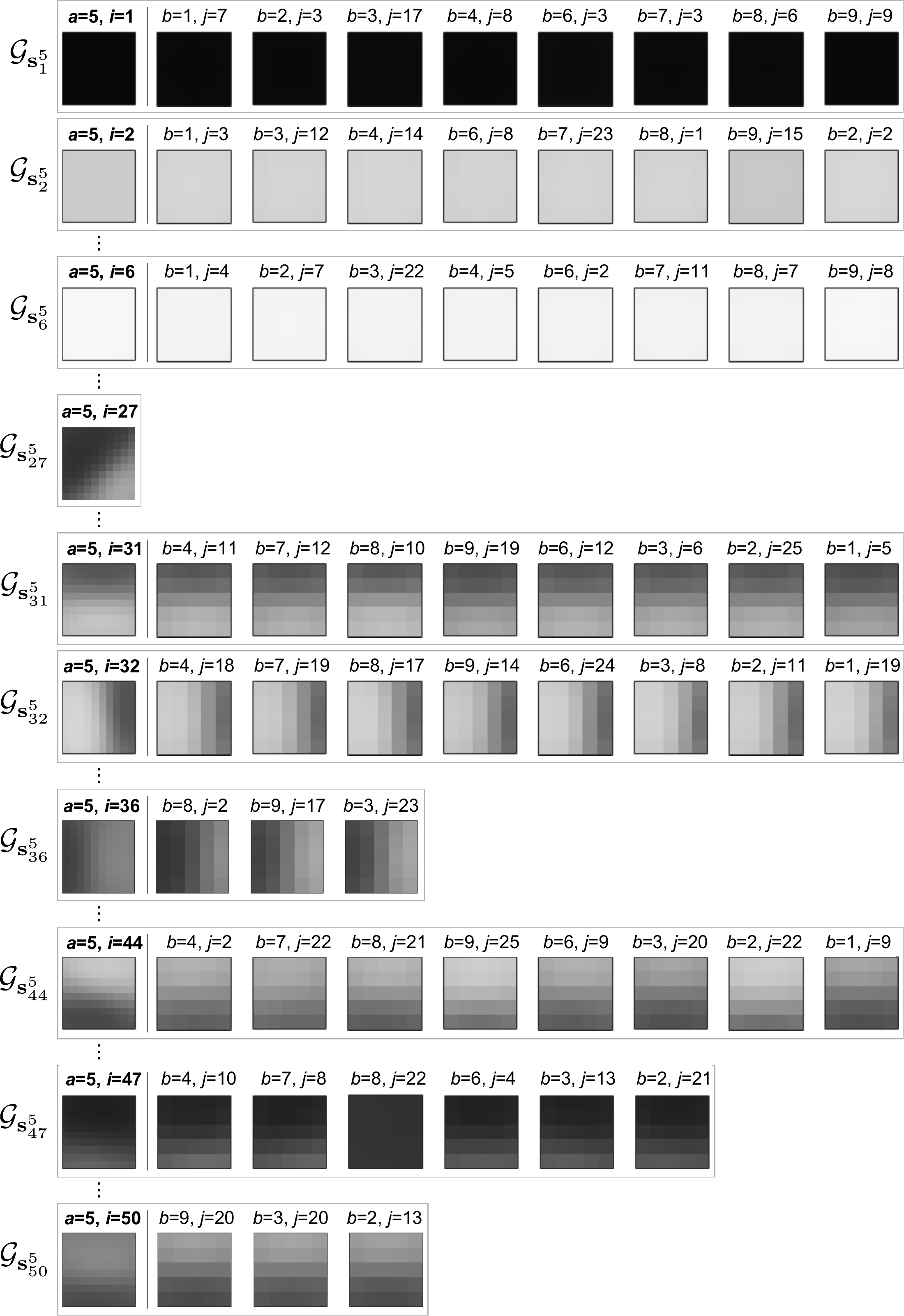}
\caption{Sets of sensory states $\mathbf{s}^b_j$ estimated as similar to foveal sensory states $\mathbf{s}^a_i$ for a threshold value $\epsilon=0.5$. Globally, the agent has been able to identify sensory states that encode the same visual features in the different receptive fields. Some sets appear incomplete due to the different clustering of sensory inputs in the receptive fields.
}
\label{fig:Sensory_groups}
\end{figure}

\textit{Visualizing similar sensory states:}
Due to the low probabilities observed in matrix $T$ when exploring random environments (see Sec.~\ref{sec:Random images}), trying to identify \emph{similar} sensory states was not a very meaningful endeavor. The situation is however different with natural images as K-Means clusters are better aligned and lead to more predictability in the model.
According to Eq.~\ref{eq:similarity_set}, we estimate sets $\mathcal{G}_{\mathbf{s}^a_i}$ of sensory states $\mathbf{s}^b_j$ considered similar to a given sensory state $\mathbf{s}^a_i$. The similarity threshold is arbitrarily set to $\epsilon=0.5$. We also limit the analysis to the foveal sensory states ($a=5$) as the central receptive field has been the only one to be shifted to all other receptive fields during the simulation. Ten of the $N^5=50$ sets $\mathcal{G}_{\mathbf{s}^5_i}$ are displayed in Fig.~\ref{fig:Sensory_groups}. One can see that, although the agent had no a priori knowledge about the correspondence between the sensory states of its different receptive fields, it has been able to group together sensory states that encode the same visual features. For $6$ of the sets $\mathcal{G}_{\mathbf{s}^5_i}$ ($i=1,2,6,31,32,44$), a similar sensory state has correctly been identified in all the other receptive fields $b$. For $3$ other sets ($i=36,47,50$), between $3$ and $6$ similar states have been identified, and no similar state was found for $i=27$. These incomplete results are due to two factors. First, the number of foveal sensory states $N^{a}=50$ is greater than the number of peripheral ones $N^a=20$. This finer covering of the foveal sensory space means that foveal sensory states do not necessarily have an equivalent in all peripheral receptive fields. Second, because K-Means has been run independently for all receptive fields, similar sensory states do not necessarily exist in all of them. This is for instance well illustrated by the presence of $\mathbf{s}^8_{22}$ in the set $\mathcal{G}_{\mathbf{s}^5_{47}}$. Visually, its pattern does not look like its counterparts but it still appears in the set because the receptive field $b=8$ does not have a better corresponding prototype. All sensory inputs looking like $\mathbf{s}^5_{47}$ are thus encoded by $\mathbf{s}^8_{22}$, the closest prototype, when encoded in receptive field $b=8$. This leads to a predictable sensorimotor transition between the two prototypes and to the naive agent evaluating them as similar.

%%%%%%%%%%%%%%%%%%%%%%%%%%%%%%%%%%%%%%%%%%%%%%%%%%%%%%%%%%%%%%%%%%%%%%%%%%%%%%%%%%%%%%%%%%%%%%%%%%
%-------------------------------------------------------------------------------------------------
%%%%%%%%%%%%%%%%%%%%%%%%%%%%%%%%%%%%%%%%%%%%%%%%%%%%%%%%%%%%%%%%%%%%%%%%%%%%%%%%%%%%%%%%%%%%%%%%%%

\section{Conclusion}
\label{sec:Discussion}

We propose a sensorimotor approach of perception, inspired by the sensorimotor contingencies theory, which claims that a naive agent should learn to master the way it can transform its sensorimotor experience. In this work, we applied such a view to explain how a robot without any a priori knowledge about the structure of its body or the structure of the world can discover the existence of the visual field generated by its unknown visual sensor. Discovering this visual field means for the agent to capture the sensorimotor regularities that it induces: different sensory states encode the same visual feature, depending on where it is encoded on the sensor array, and the agent can actively transform one into the other by sending motor commands. Those sensorimotor experiences seen from the agent's internal point of view correspond externally to the fact that visual features from the environment shift on the retina when the sensor moves and consequently get encoded by different sensels. This encoding variability is particularly marked in a heterogeneous sensor array like the human retina.

We proposed a simple visual system inspired by the retina, as well as a computational model to explain how an agent can explore and capture those sensorimotor regularities. The model is based on the definition of receptive fields, small neighborhoods of sensels that each encode only a small part of the visual scene the agent can observe. Apart from its biological motivation, the introduction of those receptive fields can be seen as an attempt to create a compositional structure in the sensor that mirrors the compositional structure of the world \cite{minsky1975framework}.
For the naive agent, all receptive fields act as independent sensors which generate their own sensory inputs in parallel. However, by actively exploring visual scenes, it can discover that its sensory experiences change in regular ways. In particular, the sensory state of some receptive fields is useful to predict what the next sensory state will be in another receptive field when it performs a certain motor command. Also, different pairs of receptive fields are coupled this way when different motor commands are sent.
One could ask why it is so important for the agent to learn and model such a thing. The answer is that this agent, that initially did not have any knowledge, can now make use of what we would call its visual field. He does know that different sensory states it experiences, and that belong to a set $\mathcal{G}_{\mathbf{s}^a_i}$, can be interpreted as similar, up to a motor command. This knowledge can be seen as the internal emergence of the concept of \emph{visual feature}, an external reality that was not previously accessible to the agent. Moreover, it also knows how to actively transform one of those sensory states into the other. This active aspect of the model is fundamental in our approach. It means that the sensory regularities extracted by the agent only make sense through action. The knowledge acquired through sensory information is thus directly able to guide action, which is the fundamental role of \emph{perception}. The agent could for instance use its estimated model to perform a visual search task. Let us assume that, for some reason, receiving the sensory state $\mathbf{s}^5_{18}$ ($18^\text{th}$ state of the foveal receptive field $5$) is rewarding for the agent. Given its internal model, it can counterfactually \cite{seth2014predictive} determine if a sensory state from another receptive field does correspond to the same visual feature and which saccade to execute to reach the rewarding sensory state. This is the kind of visual tasks that were proposed to subjects with altered sensory states associations in \cite{herwig2014predicting} and for which a basic model has recently been proposed in \cite{laflaquiere2016ICDL}.

This use of the predictive model has not been illustrated in this paper. It is part of the multiple future directions in which we intend to extend this work.
First, the system proposed in the simulation was very simple compared to its biological counterpart. It will be replaced by a larger retina split into more receptive fields organized in multiple concentric rings with decreasing visual resolutions. A parallel version of the algorithm used in this simulation will however be necessary to benefit from the computational power of a GPU and process efficiently the huge amount of parallel data that such a richer sensor will generate. This technical improvement can be seen as a computational requirement but also as a direct inspiration from our visual nervous system.
Another direction of future research is the analysis of the environmental structure captured in the predictive model. As mentioned in Sec.~\ref{sec:Natural images}, this structure appears in particular in unpaired blocks and reveals properties of the environment. According to the SMCT, it is thus possible to cluster sensory states that encode the same visual features, but it is also possible to assess properties of those features based on the way they get transformed by actions. A simple example proposed in \cite{o2001sensorimotor} is the one of an horizontal line. Regardless of the way such a visual feature is encoded by the sensor, it is true that this sensory input will be invariant to a leftward or rightward movement of the sensor. Characterizing visual features will also provide a better picture of the experience of \emph{seeing} in the sensorimotor approach of perception by describing the intermingled influence of the sensor and environmental structures. 
The predictive model used in the simulation should also be modified to improve the predictability between sensory states that encode the same visual features. The current model is far from satisfying: it discards a lot of information by discretizing the sensory spaces and relies on the alignment of clusters in sensory spaces that do not necessarily share the same properties. A more powerful approach, but also more complex to train, would be to directly estimate a non-linear mapping between the different sensory spaces associated with the receptive fields. Each sensory data could be processed through those mappings without relying on clusters and the associated loss of information. This would lead to a more satisfying notion of equivalence between sensory states by increasing the threshold value $\epsilon$ (see Sec.~\ref{sec:Formalization and learning of the predictive model}).
Finally, the most challenging improvement will be to take into account continuous motor commands instead of the very limited discrete set of saccades considered in the current simulation. Continuous saccades might require a less constrained definition of receptive field for which any sensel and its neighbors could be considered a receptive field, leading to an almost continuous coverage of the retina by overlapping receptive fields.

Despite its current limitations, the model introduced in this paper proposes an innovative perspective to address the problem of visual perception in artificial systems. It relies on the division of the sensor into smaller sensory units and on extracting regularities in the way actions transform their states. This is very different from traditional approaches which generally process static images as a whole to extract interesting features. Of course, many algorithms, such as convolutional neural networks, do focus on small patches in the image that could be compared to the receptive fields used in this work. However, they do so based on the strong hypothesis that all patches share the same properties and encode visual information the same way. This assumption is not realistic for a heterogeneous sensor such as the human eye. Our approach is more general and is able to deal with heterogeneous sensors. It also emphasizes the fact that action is a necessary component for a naive agent to extract useful information from an uninterpreted sensory flow. In fact, the algorithm used in this work is general enough that it could process information coming from different kinds of sensor arrays. For instance, the simulated sensor described in this paper could just as much be a tactile sensor and the visual scenes could be tactile textures that the agent would touch. The agent would discover properties of the ``field of touch" the sensor generates and how tactile features move in the array when motor commands are sent.
Moreover, the sensorimotor approach also seems promising to address problems related to multimodality. Without a priori knowledge, one does not need to assume the existence of separate modalities. The agent would thus naturally extract sensorimotor regularities that combine multiple modalities.
Yet, an even more interesting and challenging question will be the one related to the co-discovery of contingencies by a naive agent. So far the sensorimotor approach of perception has been applied to capture regularities associated with targeted properties of the world thanks to specific settings: space, colors, environments, objects, and now visual field \cite{laflaquiere2015learning, terekhov2013space, philipona2006color, witzel2015, laflaquiere2015grounding, laflaquiere2015objects}. The next difficult problem is to develop a system that is able to extract those different contingencies, and others, in parallel. This should reveal the intrinsic intermingling of the notion of visual field with the one of space, especially when extending the set of actions to $3$D displacements in the environment.

\bibliographystyle{elsarticle-harv}
\bibliography{bib_short}

%\begin{thebibliography}{99}

%\bibitem{c1} [1]J. O'Regan and A. Noë, 'A sensorimotor account of vision and visual consciousness', Behavioral and Brain Sciences, vol. 24, no. 05, pp. 939-973, 2001.
%\end{thebibliography}

\end{document}